%% file: paper.tex
\documentclass[runningheads]{llncs}
\usepackage[T1]{fontenc}
\usepackage{graphicx}
\usepackage[utf8]{inputenc} 
\usepackage[T1]{fontenc}    
\usepackage{hyperref}       
\usepackage{url}            
\usepackage{booktabs}       
\usepackage{amsfonts}       
\usepackage{nicefrac}       
\usepackage{microtype}      
\usepackage{xcolor}  
\usepackage{tabularx}

\begin{document}
\title{Kwame 2.0: Human-in-the-Loop Generative AI Teaching Assistant for Large Scale Online Coding Education in Africa}
\titlerunning{Kwame 2.0}
%

\author{George Boateng\inst{1,3} \orcidID{0000-0003-4540-9582} \and
Samuel Boateng\inst{3}\orcidID{0009-0005-9809-1290}\and
Victor Kumbol \inst{2,3}  \orcidID{0000-0002-6828-5570}}
\authorrunning{G. Boateng et al.}

\institute{ETH Zurich, Switzerland\\
\and
Charité - Universitätsmedizin Berlin, Germany\\
\and
Kwame AI Inc., U.S.}



\maketitle 


\begin{abstract}
Providing timely and accurate learning support in large-scale online coding courses is challenging, particularly in resource-constrained contexts. We present Kwame 2.0, a bilingual (English–French) generative AI teaching assistant built using retrieval-augmented generation and deployed in a human-in-the-loop forum within SuaCode, an introductory mobile-based coding course for learners across Africa. Kwame 2.0 retrieves relevant course materials and generates context-aware responses while encouraging human oversight and community participation. We deployed the system in a 15-month longitudinal study spanning 15 cohorts with 3,717 enrollments across 35 African countries. Evaluation using community feedback and expert ratings shows that Kwame 2.0 provided high-quality and timely support, achieving high accuracy on curriculum-related questions, while human facilitators and peers effectively mitigated errors, particularly for administrative queries. Our findings demonstrate that human-in-the-loop generative AI systems can combine the scalability and speed of AI with the reliability of human support, offering an effective approach to learning assistance for underrepresented populations in resource-constrained settings at scale.

\end{abstract}

\input{content.tex}


\bibliographystyle{splncs04}
\bibliography{refs}

\end{document}

%% file: content.tex
\section{Introduction}
Learners in introductory coding courses often require timely, accurate support to overcome conceptual challenges and sustain engagement. In traditional face-to-face settings, human teaching assistants and instructors fill this role; however, scaling such support for large online cohorts presents significant challenges, especially in under-resourced environments. To address such concerns, in 2020, Boateng \cite{boateng2021b} developed Kwame, a bilingual AI teaching assistant that provided passages from lesson notes to answer students' questions in both English and French, within SuaCode, a smartphone-based introductory online coding course for Africans. However, Kwame operated as a static retrieval system and was not deployed for real-time learning support.

Recent advances in generative artificial intelligence (AI) — particularly large language models (LLMs) — provide new opportunities to offer timely learning support at scale in coding courses \cite{raihan2025}. We aimed to extend Kwame with generative AI capabilities. Consequently, we built Kwame 2.0, a retrieval-augmented generation (RAG) \cite{lewis2020} system that allows it to deliver more context-aware and effective responses in English and French, thereby enhancing the quality of support learners receive in real-time. Unlike generic chatbot interfaces, we deployed it in a human-in-the-loop setup that addressed student questions with AI and community support from both learners and facilitators. We deployed Kwame 2.0 in a large-scale longitudinal study of 15 monthly cohorts from October 2024 to December 2025 with over 3,717 enrollments from learners across 35 of the 54 African countries. We evaluated Kwame 2.0’s (1) helpfulness using ratings from the community and (2) accuracy using expert ratings.

Our key contributions are (1) the design and deployment of a bilingual generative AI teaching assistant tailored for coding education, (2) an evaluation of its effectiveness and limitations in real-world learning environments in a large-scale longitudinal study, and (3) evidence that human-in-the-loop AI support can combine the speed and scalability of generative AI with the accuracy and contextual judgment of humans — yielding improved learning support for underrepresented learners at scale in resource-constrained settings.

\section{Background and Related Work}
In this section, we describe background work on SuaCode and Kwame, and related work on using generative AI for learning support in coding education.

\subsection{SuaCode}
SuaCode \footnote{\href{http://suacode.ai/}{http://suacode.ai/}} is an AI-powered smartphone app (Android) that provides online coding courses with lesson notes, exercises, quizzes, and fun coding assignments in English and French, which cover official languages across Anglophone and Francophone Africa. It runs as monthly cohorts. The SuaCode builds upon prior annual cohort runs of a smartphone-based introductory coding course for Africans in-person in 2017 \cite{boateng2018} and online from 2018 to 2020 \cite{boateng2019,boateng2021,suacodeafrica2}. It uses lesson notes designed for offline reading to reduce Internet data usage since data is expensive across Africa \cite{dw2020}. The courses adopt a project-based learning approach where learners build and interact with a game on their phones as assignments. Each section of the course ends with multiple-choice quizzes and a coding assignment. The assignments are graded by an automated grading software, AutoGrad \cite{annor2021}, which provides detailed individualized explanations for wrong answers with an option for learners to submit a complaint in cases where they disagree with the provided answers and feedback. It also provides a cohort-specific, AI-powered, human-in-the-loop forum that is designed to allow learners to ask questions (even anonymously) and get quick and accurate answers from their peers, facilitators, and an AI teaching assistant, Kwame \cite{boateng2021b}. Kwame enables learners to get individualized learning support so they do not quit when it gets tough. Each person’s name shows an earned badge to encourage helpful engagement. Each of the 3 badges (bronze, silver, and gold) is received based on a “helpfulness score” that we calculate using various metrics such as upvotes on questions and answers contributed by each learner. It has a leaderboard that celebrates helpful learners and facilitators, thereby encouraging everyone in the cohort to be helpful. Learners who complete receive certificates and mentoring. 

\subsection{Kwame}
Kwame \cite{boateng2021b} is a bilingual AI teaching assistant that was developed for SuaCode courses. It is a retrieval-based question-answering system that was trained using the SuaCode course material (lesson notes, quizzes, past questions, and answers) and evaluated using accuracy and time to provide answers. When it receives a question, it first detects if it’s an English or French question, and then it finds the paragraph among the bank of paragraphs from the lesson notes in the detected language that is most semantically similar to the question via cosine similarity with a fine-tuned Sentence-BERT \cite{reimers2019} model and returns it along with the similarity score. Evaluations of Kwame and comparison with similar assistants from prior work showed superior accuracy and fast response time. Kwame was never deployed in a real course setting and only had offline accuracy and latency evaluations using prior SuaCode course content.

\subsection{Generative AI for Coding Education Support}
Several works, such as CS50 Duck \cite{liu2024}, CourseAssist \cite{feng2024}, CodeTutor \cite{lyu2024}, and CodeHelp \cite{liffiton2023} have developed generative AI tools and deployed them to provide learning support in coding education through question answering. These tools have used RAG designs, and their evaluations have demonstrated utility in terms of the support received by learners in their coding course. These works have not comprehensively evaluated these tools in human-in-the-loop settings.  Our work uses similar approaches, such as RAG. The key way our work differs is its deployment in a large-scale online coding course in a (1) human-in-a-loop setting where the AI is interacted with in a forum rather than as a chatbot and (2) a learning context targeting an underrepresented demographic in a resource-constrained setting — learners across Africa in a smartphone-based coding course with limited teaching assistants or facilitators.

\section{System Overview}
\begin{figure}[t]
  \centering
  \includegraphics[width=\linewidth]{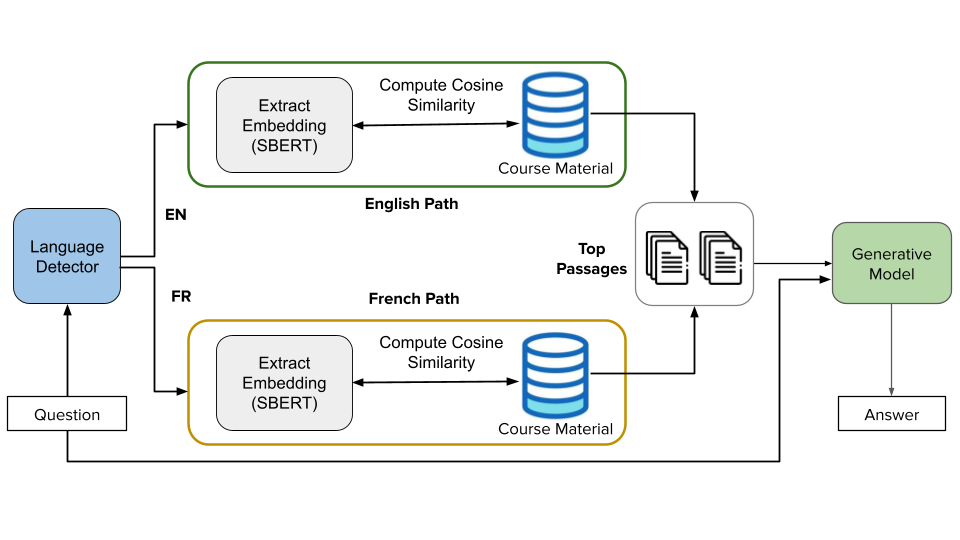}
  \caption{Architecture of Kwame 2.0}
  \label{fig:kwame_architecture}
\end{figure}

\begin{figure}[t]
  \centering
  \includegraphics[width=\linewidth]{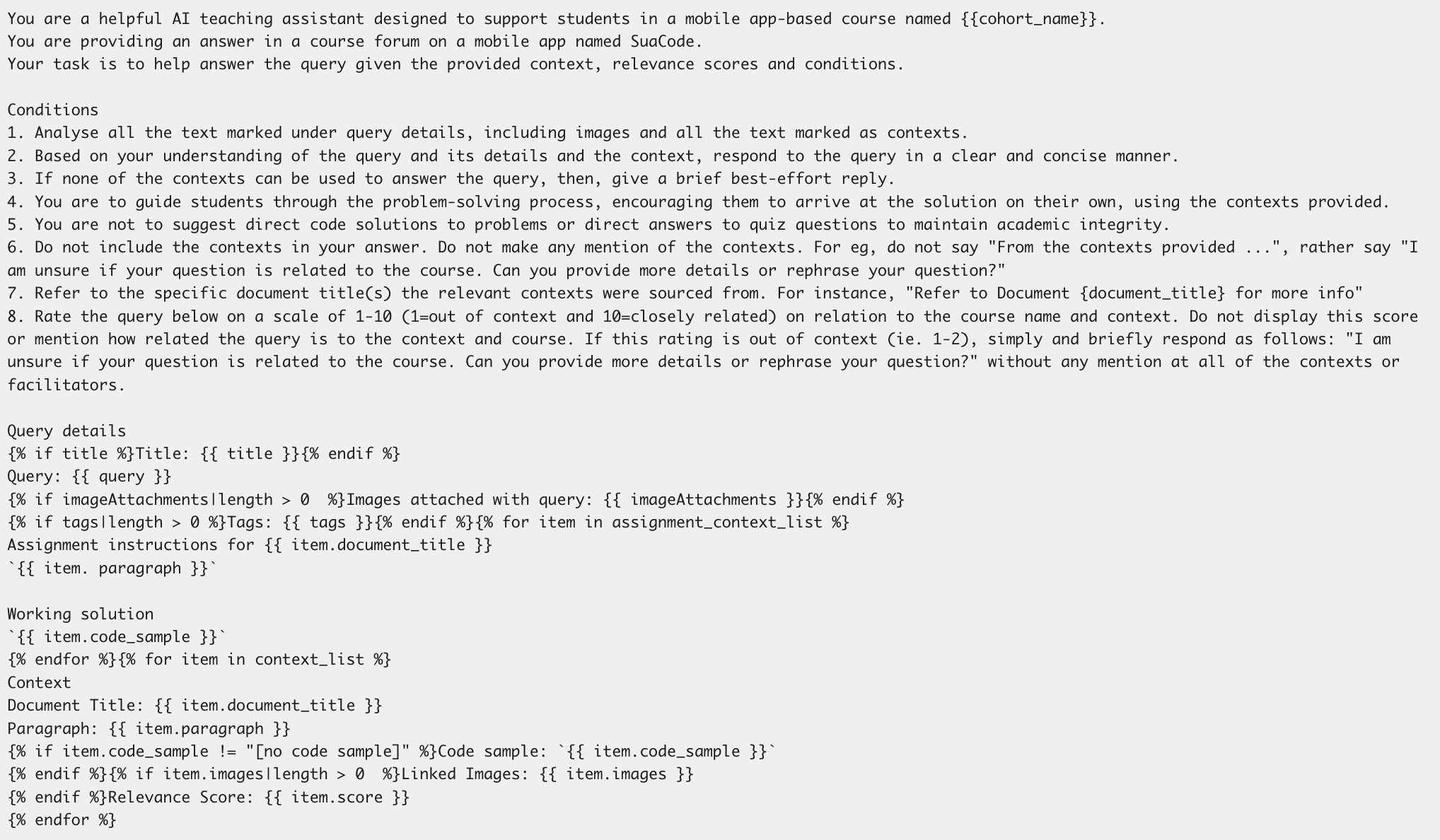}
  \caption{Prompt for Kwame 2.0}
  \label{fig:kwame_prompt}
\end{figure}

Kwame 2.0 is a RAG system that uses the retrieval pipeline of Kwame 1.0 (Figure \ref{fig:kwame_architecture}). When question is asked, our system detects the language of the question, computes an embedding of the question using Sentence-BERT, and then computes cosine similarity scores with all saved embeddings in ElasticSearch to retrieve the top 5 relevant passages from our course materials based on students’ questions, any attached images, and code snippets from relevant course material using question tags, and then uses them as context to the GPT-4 API \cite{achiam2023} to generate a direct answer almost instantly. We developed a well-engineered prompt (Figure \ref{fig:kwame_prompt}) that only provided hints and suggestions rather than whole-code solutions to students’ questions about assignments. Also, Kwame 2.0 provided personalized responses to students when they introduced themselves in the course forum, though it was not specifically designed to do so. To address concerns of hallucinations by the model, all answers are based on retrieved passages from the course materials, which are cited at the end of each answer as evidence and resources, to encourage further reading. 

\section{Deployment}
\begin{figure}[t]
  \centering
  \includegraphics[width=0.5\linewidth]{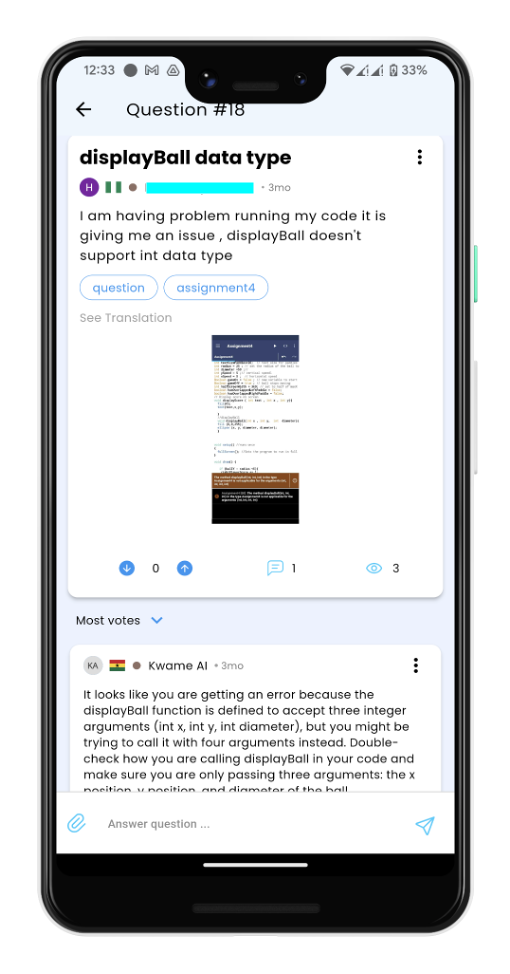}
  \caption{Screenshots of Kwame 2.0 in the SuaCode course forum}
  \label{fig:kwame_forum}
\end{figure}

We deployed Kwame 2.0 as a “facilitator” (Figure \ref{fig:kwame_forum}) that provided one answer to each question in the cohort-specific public forum of the “Introduction to Programming” course in English and French via the SuaCode app in a longitudinal study of 15 monthly cohorts from October 2024 to December 2025. Learners did not pay for the course and received scholarships due to a partnership with Deloitte. There were a total of 3,717 enrollments by learners (average age was 26.2 years, std of 8.4) across 35 of the 54 African countries. Each cohort ran as a structured, human-facilitated learning environment to ensure that AI responses were complemented by human oversight. In particular, we had 2 facilitators along with learners who could respond to queries, upvote, or downvote answers. Furthermore, a user who asked a question could accept only one answer as the correct one. This human-in-the-loop model ensured that Kwame 2.0’s responses were adequately supported and supervised.

\section{Evaluation}
We evaluated the helpfulness of Kwame 2.0’s answers using community ratings (i.e., the number of upvotes or downvotes on answers in the forum by those in the cohort) and the number of questions that had an answer being accepted as correct in the forum. We also had expert ratings by one of the course facilitators — creator of the course — who categorized the questions as valid or invalid (i.e, required an answer), curricular or administrative questions, and rated each answer from Kwame as correct or incorrect. Given the iterative development of Kwame 2.0 and the vastness of answers, we evaluated the last quarter of five quarters of SuaCode cohorts (October to December 2025), which contained 536 questions.

Overall, learners and facilitators provided little feedback on answer quality in the forum and were slow to provide answers to questions. Only 24 questions had an accepted answer, with 83.3\% being answers from Kwame 2.0 and 16.7\% from the community. Also, 36 questions had at least one answer that was upvoted with 69\% from Kwame 2.0 and 41.7\% from the community (note that a question could have upvoted answers from both Kwame 2.0 and the community). These show that Kwame 2.0’s answers were of high quality. Given that Kwame 2.0’s answers were provided almost instantaneously, they had a good chance to address learners’ questions early and then be accepted as the correct answer and or upvoted.

Our expert evaluation showed that there were 490 valid questions, out of which 50.8\% were curriculum questions and 49.2\% were administrative questions. We computed accuracy as the number of correct answers from Kwame 2.0 for valid questions. Kwame 2.0 had an accuracy of 76.7\% with 97.6\% for curriculum questions, and 46.9\% for administrative questions. Analysis of the incorrect answers showed that they resulted from gaps in information in the course material, especially for administrative questions. This gap can be addressed simply by having up-to-date administrative information indexed. It is important to note that even incorrect answers usually contained helpful information, such as referring students to specific course materials or asking them to contact course instructors. Additionally, facilitators and other students jumped in to provide correct answers, which made up for the incorrect answers by Kwame 2.0. Of the 114 questions for which Kwame 2.0 did not provide a correct answer, 38.6\% were subsequently answered correctly by either another student or a facilitator, increasing the overall accuracy to 85.7\% when combining AI and community responses. This human-in-the-loop setup allowed learners to receive fast and generally accurate responses from Kwame 2.0, while ensuring that incorrect or missing AI answers were later corrected by peers or facilitators. These findings provide strong evidence for deploying AI assistants in human-in-the-loop configurations, which effectively combine the speed and scalability of AI with the oversight — albeit greater latency — of human support, a critical consideration in high-stakes learning contexts.

\section{Conclusion}
In this work, we developed Kwame 2.0, a bilingual generative AI teaching assistant deployed in a large-scale online introductory coding course for learners across Africa. We built it using retrieval-augmented generation and deployed it in a human-in-the-loop forum, which provided timely learning support grounded in course materials and community oversight. Results from a 15-month longitudinal deployment with over 3.7K enrollments showed that Kwame 2.0 achieved high accuracy on curriculum-related questions and that human facilitation effectively compensated for its limitations, particularly on administrative queries. The combined AI-plus-community responses significantly improved overall accuracy, demonstrating the value of human-in-the-loop designs in educational settings. These findings highlight the potential of generative AI to scale learning support when thoughtfully integrated with human expertise, especially for underrepresented learner populations in resource-constrained contexts. Future work will evaluate users' qualitative feedback on Kwame 2.0 and assess its impact on learning outcomes.

\begin{credits}
\subsubsection{\ackname} We are also grateful to ETH for Development (ETH4D), whose funding enabled the development of the SuaCode platform. We are also grateful to Deloitte Switzerland, whose funding allowed us to provide scholarships to learners to take SuaCode's “Introduction to Programming” course and also enabled the implementation of Kwame 2.0. 
\end{credits}